\documentclass[letterpaper, 10 pt, conference]{ieeeconf}  %
\IEEEoverridecommandlockouts
\usepackage{etex}
\usepackage{cite}
\usepackage{amsmath,amssymb,amsfonts}
\usepackage{graphicx}
\usepackage{textcomp}
\usepackage{xcolor}
\def\BibTeX{{\rm B\kern-.05em{\sc i\kern-.025em b}\kern-.08em
    T\kern-.1667em\lower.7ex\hbox{E}\kern-.125emX}}

\usepackage{url} 

\usepackage[T1]{fontenc}
\usepackage[utf8]{inputenc}

\usepackage{booktabs}

\usepackage{bm}   
\usepackage{color}

\usepackage{tikz,pgfplots}
\pgfplotsset{compat=newest}
\pgfplotsset{plot coordinates/math parser=false}
\usetikzlibrary{plotmarks}

  \usepackage[ruled, vlined, nofillcomment]{algorithm2e}
  \usepackage{algorithmic}
  
  \SetCommentSty{mycommfont}
  \SetAlgoCaptionSeparator{:}

  \usepackage{calc}
\usepackage[capitalise,nameinlink]{cleveref}
\crefname{equation}{Eq.}{Eqs.}
\crefname{algorithm}{Alg.}{Algs.}

\IEEEtriggeratref{16}

\usepackage{notation}   
   
\begin{document}

\title{\LARGE \bf Online One-Dimensional Magnetic Field SLAM \\ with Loop-Closure Detection}

\author{Manon Kok$^{1}$ and Arno Solin$^{2}$%
\thanks{This publication is part of the project ``\emph{Sensor Fusion for Indoor Localisation Using the Magnetic Field}'' with project number 18213 of the research program Veni which is (partly) financed by the Dutch Research Council (NWO). Arno Solin also acknowledges funding from the Research Council of Finland (grant id 339730).
}%
\thanks{$^{1}$Manon Kok is with the Delft Center for Systems and Control, Delft University of Technology,
        the Netherlands
        {\tt\small m.kok-1@tudelft.nl}.}%
\thanks{$^{2}$Arno Solin is with the Department of Computer Science, Aalto University, Finland
        {\tt\small arno.solin@aalto.fi}.}%
}

\maketitle
\thispagestyle{empty}
\pagestyle{empty}

\begin{abstract}
We present a lightweight magnetic field simultaneous localisation and mapping (SLAM) approach for drift correction in odometry paths, where the interest is purely in the odometry and not in map building. We represent the past magnetic field readings as a one-dimensional trajectory against which the current magnetic field observations are matched. This approach boils down to sequential loop-closure detection and decision-making, based on the current pose state estimate and the magnetic field. We combine this setup with a path estimation framework using an extended Kalman smoother which fuses the odometry increments with the detected loop-closure timings. We demonstrate the practical applicability of the model with several different real-world examples from a handheld iPad moving in indoor scenes.
\end{abstract}

\section{Introduction}

Spatial variations of the magnetic field can be used as a source of position information for indoor localisation~\cite{Haverinen+Kemppainen:2009,Li+Gallagher+Dempster+Rizos:2012,Angermann+Frassl+Doniec+Julian+Robertson:2012,LeGrand+Thrun:2012,Solin+Sarkka+Kannala+Rahtu:2016,Hanley+Faustino+Zelman+Degenhardt+Bretl:2017,Gao+Harle:2017,Torres+Rambla+Montoliu+Belmonte+Huerta:2015,Kok+Solin:2018,visetHK:2022,ouyangA:2022}. These variations are typically due to ferromagnetic material in the structures of buildings and to a lesser extent due to the presence of for instance furniture (see, \eg, \cite{Kok+Solin:2018}). The advantage of using the magnetic field for localisation is that it can be measured by small devices, without additional infrastructure or line-of-sight requirements. Furthermore, magnetometers are nowadays present in (almost) any smartphone. However, accurate, online, large-scale localisation based on only the magnetic field in combination with odometry information remains an open problem.

\begin{figure}[t]
\centering
\setlength{\figurewidth}{0.87\columnwidth}
\setlength{\figureheight}{0.75\columnwidth}
\pgfplotsset{axis on top,tick label style={font=\tiny}}
\input{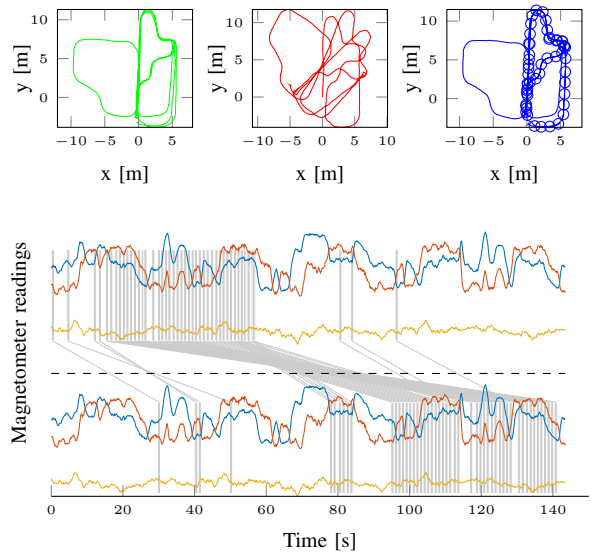}\\[-1em]
\caption{Illustration of our method with top left (green) ground truth, top middle (red) odometry, top right (blue) estimation results. The blue circles indicate the locations at which loop closures are detected. In the bottom, we twice visualise the magnetic field measurements (in a body-fixed frame) over time, above and below the black dashed line. In grey we indicate the locations where the magnetic field is detected to be similar in which we use loop closures in our algorithm.}
\label{fig:library}
\end{figure}

In this work, we present a novel approach to magnetic field simultaneous localisation and mapping (SLAM), where the interest is purely in localisation and not in the map itself. We assume that the motion is in 1D, \ie, through corridors in buildings, which leads to 1D trajectories through the ambient magnetic landscape. These types of paths are typical in the case of indoor localisation. Because of this assumption, it is not necessary to explicitly construct a (2D or 3D) map of the magnetic field as is frequently done in existing work (see, \eg, \cite{Kok+Solin:2018,Angermann+Frassl+Doniec+Julian+Robertson:2012,LeGrand+Thrun:2012,visetHK:2022,kok2024rao}). Instead, we present a heuristic approach for detecting loop closures, which makes our method significantly more lightweight than approaches that construct a magnetic field map. In detecting these loop closures we assume that the user does not rotate the mobile sensor with respect to the body during use. We use odometry information in combination with these loop closures in an extended Kalman smoother, and illustrate the applicability using real-world examples. This is illustrated in \cref{fig:library}, where in the top figures the ground truth is shown in green, the odometry in red, and the estimates from our algorithm in blue. The blue circles indicate the locations at which we identified loop closures in the algorithm. In the bottom figure, the three components of the magnetic field measurements (in a body-fixed frame) are displayed over time. We depict these measurements twice, above and below the dashed line to explicitly indicate in grey the times at which the magnetic field is determined to be similar to a previous time instance, causing loop closures to be detected and used in our algorithm.

The key contribution of this paper is in providing a novel both insightful and lightweight magnetic SLAM approach. We demonstrate the practicality of the proposed approach in several real-world proof-of-concept data sets collected with consumer-hardware. A reference implementation of our method and code to reproduce the experiments can be found on \url{https://github.com/manonkok/1d-magnetic-field-slam/}.

This paper uses a fundamentally different approach than our existing work~\cite{Kok+Solin:2018,visetHK:2022,osmanVK:2022} in which we studied 3D magnetic field SLAM and constructed maps of the magnetic field using Gaussian processes~\cite{Rasmussen+Williams:2006}. We overcome the challenge that constructing these Gaussian process-based magnetic field maps is computationally expensive by restricting our interest to one-dimensional paths which allows for using explicit loop closures using magnetic field features. One-dimensional magnetic field SLAM has previously been studied in \cite{Siebler+Lehner+Sand+Hanebeck:2024} for train localization, where the railway tracks make the motion inherently one-dimensional. For indoor localization, one-dimensional magnetic field localisation has previously been studied in~\cite{suksakulchaiTWK:2000} for a mobile robot traversing corridors. Contrary to our approach, their focus is only on localisation and not on SLAM. Magnetic field features have been previously detected and classified both directly based on their sensor properties and using machine learning, see \eg\ \cite{subbuGD:2011,Montoliu+Torres-Sospedra+Belmonte:2016,leeAH:2018,carrilloMUS:2015,wangJDCWHA:2024}.
Contrary to these methods, our algorithm does not rely on any previously available magnetic field measurements and recursively detects if a location is re-visited in the current data set. 

\section{Modelling}
\label{sec:modelling}
We are interested in jointly estimating a time-varying, two-dimensional sensor position $\vp$, a time-varying sensor heading~$\psi$ and a number of stationary magnetic landmarks~$\vl_k, k = 1,2, \hdots, K$. We express the sensor position $\vp$ in a \emph{world frame} $\wframe$ and the heading as the rotation from a \emph{body frame} $\bframe$ to the world frame. We choose the origin of the world frame to be equal to the initial position of the sensor. Without loss of generality we assume the axes of the world frame to be aligned with the gravity and the initial heading. Furthermore, we assume that the inclination of the sensor is known, \eg~from accelerometer measurements. Hence, we define the inclination of the body frame to be aligned with the gravity and the heading of the body frame to be aligned with the sensor axes. In this section, we present the state-space model that we use in \cref{sec:inference} to estimate the position and heading of the mobile platform.

\subsection{Dynamic model}
\label{sec:dynModel}
We assume that odometry measurements $\vy_{\text{p},t}$ and $y_{\omega,t}$ are available that provide information about the change in position $\Delta \vp$ and the change in heading $\Delta \psi$, respectively. More specifically, we model the measurements $\vy_{\text{p},t} \in \mathbb{R}^2$ as 
\begin{equation}
\vy_{\text{p},t} = \Delta \vp_t^\bframe + \ve_{\text{p},t}, \qquad \ve_{\text{p},t} \sim \mathcal{N}(0, \sigma_\text{p}^2 \boldsymbol{\mathcal{I}}_2).
\end{equation}
Note that these measurements are expressed in the body frame~$\bframe$, \ie, they are aligned with the gravity and the sensor axes. These measurements can for instance be obtained using wheel encoders, using pedestrian dead-reckoning algorithms or using odometry measurements provided by any other sensor set. The measurements $y_{\omega,t} \in \mathbb{R}$ providing information about the change in heading $\Delta \psi$ are in practice typically obtained using gyroscopes. We model these measurements as 
\begin{equation}
y_{\omega,t} = \omega_t + b_\omega + e_{\omega,t}, \qquad e_{\omega,t} \sim \mathcal{N}(0, \sigma^2_\omega),
\end{equation}
where $\omega_t$ refers to the angular velocity around the gravity direction. To model sensor imperfections in the gyroscopes, we model that these measurements are corrupted by Gaussian i.i.d.\ noise $e_{\omega,t}$ and by a constant sensor bias~$b_\omega$.

We therefore define our state vector to not only consist of the time-varying sensor position $\vp$, the time-varying sensor heading~$\psi$, and $K$ stationary magnetic landmarks $\vl_k, k = 1,2, \hdots, K$ but also of a constant sensor bias $b_\omega$. Our dynamic model uses the odometry measurements as an input as 
\begin{subequations}
\begin{align}
\vp_{t+1}^\wframe &= \vp_t^\wframe + \MR(\psi_t) \left( \vy_{\vp,t} - \ve_{\text{p},t} \right), \\
\psi_{t+1} &= \psi_t +  T_t \left( y_{\omega,t} - \bias{t} - e_{\omega,t} \right), \\
\bias{t+1} &= \bias{t}, \\
\vl_{k, t+1} &= \vl_{k,t}, \label{eq:dynModel-landmark}
\end{align}
\label{eq:dynModel}%
\end{subequations}%
where $T_t$ denotes the sampling time at time $t$. Furthermore, $\MR(\psi_t) = \begin{pmatrix} \cos \psi_t & \sin \psi_t \\ - \sin \psi_t & \cos \psi_t \end{pmatrix}$ denotes the rotation matrix from the body frame to the world frame. Note that $\MR\T(\psi_t)$ denotes the rotation matrix from the world frame to the body frame.

\subsection{Measurement model}
We assume that the only additional information on top of the dynamic model from \cref{sec:dynModel} is information obtained by detection of magnetic landmarks. When the magnetic field at time instance $t$ is determined to be equivalent to the magnetic field at time instance $t - \ell$, we introduce a loop closure. For the $k$\textsuperscript{th} loop closure we now model the position $\vp_t^\wframe$ and the position $\vp_{t-\ell}^\wframe$ to both be approximately equal to the location~$\vl_{k}$ of the $k$'th landmark as
\begin{equation}
\vp_t^\wframe =  \vl_{k,t} + \ve_{\text{lc},t}, \qquad \vp_{t-\ell}^\wframe =  \vl_{k,t-\ell} + \ve_{\text{lc},t-\ell},
\label{eq:measModel}
\end{equation}
where $\ve_{\text{lc},t}$ is i.i.d.\ with $\ve_{\text{lc},t} \sim \N(0,\sigma_\text{lc}^2 \boldsymbol{\mathcal{I}}_2)$. Note that \cref{eq:measModel} enforces a loop closure because the landmark locations are assumed to be constant, see \cref{eq:dynModel-landmark}, and hence $\vl_{k,t} = \vl_{k,t-\ell}$. The reason to nevertheless distinguish between $\vl_{k,t}$ and $\vl_{k,t-\ell}$ is to retain the Markov property of our system. 

\section{Detecting landmarks}
\label{sec:detectionLandmarks}
To detect loop closures, we detect at which time instances the magnetic field is similar to the field observed previously. To this end, we assume the availability of magnetic field measurements $\vy_{\text{m},t}^\text{b}$ in body frame. Note that in practice magnetometers provide measurements in a frame aligned with the sensor axes. However, as discussed in \cref{sec:modelling}, we assume that the inclination is known and use this to rotate the magnetometer measurements to the gravity-aligned body frame.\looseness-1

Since we assume that the user either walks the same path twice or traverses the path in the opposite direction and we furthermore assume that the relative orientation of the sensor with respect to the user is constant over time, we can directly use the magnetic field measurements $\vy_{\text{m},t}^\text{b}$ for detecting loop closures. Since the magnetic field is only a three-dimensional vector and it is therefore not unique at different locations, we use a number of subsequent measurements to detect loop closures. At time instance $t$, we therefore compute two sets of weights, $\weightm{t,i}^\text{fwd}$ and $\weightm{t,i}^\text{bwd}$, as
\begin{align}
\weightm{t,i}^\text{fwd} &{=} \prod_{n = 0}^{N_\text{lc}-1} \!\! \exp\! \left( - \tfrac{1}{12 \sigma_\text{m}^2} \| \vy_{\text{m},i-n}^\text{b} - \vy_{\text{m},t-n}^\text{b} \|_2^2  \right), \label{eq:magWeights} \\
\weightm{t,i}^\text{bwd} &{=} \prod_{n = 0}^{N_\text{lc}-1} \!\! \exp\! \left( - \tfrac{1}{12 \sigma_\text{m}^2} \| \vy_{\text{m},i+N_\text{lc}-1-n}^\text{b} - \widetilde{\MR}(\pi) \vy_{\text{m},t-n}^\text{b} \|_2^2  \right), \nonumber 
\end{align}
which represent how likely loop closures in the forward and backward direction are, respectively. Here, $\widetilde{\MR}(\pi)$ is a $3 \times 3$ matrix that rotates the measurements $180^\circ$ around the $z$-axis and $N_\text{lc}$ is a tuning parameter. Note that~\cref{eq:magWeights} is directly inspired by the fact that we assume that the magnetometer measurement noise is i.i.d.\ and $\ve_{\text{m},t} \sim \N(0,\sigma^2_\text{m} \mathcal{I}_3)$, with the factor 12 arising from the assumption of the independence of the measurements and the fact that these are three-dimensional. The larger the weight $\weightm{t,i}^{\text{fwd}}$ or $\weightm{t,i}^{\text{bwd}}$ is, the more likely it is that the location at time $t$ is equal to that of time $i$.

Since we detect magnetic landmarks while estimating the position $\vp$, the estimated position $\hat{\vp}_t$ and its corresponding covariance $P_{\text{p},t}$ provide additional information about the probability that the position $\vp_t$ is equal to a previous position. We therefore compute a second set of weights
\begin{equation}
\weightp{t,i} = \exp \big( - \tfrac{1}{8 \sigma_{\text{w}_\text{p}}^2} \| \hat{\vp}_t - \hat{\vp}_i  \|_2^2  \big), \qquad t < i,
\label{eq:posWeights}
\end{equation}
where $\sigma_{\text{w}_\text{p}}$ is the mean of the square root of the diagonal elements of $P_{\text{p},t}$. Again, \cref{eq:posWeights} is inspired by the fact that we approximate the estimation error to be Gaussian. To simplify computation and keep our implementation light-weight, in the computation of the weights in \cref{eq:posWeights} we approximate the estimation errors of $\hat{\vp}_t$ and $\hat{\vp}_i$ to be independent of each other and their individual components to be independent. Note that this is not a severe approximation, as this is only used as a heuristic to computing when loop closures occur. 

The overall weights $w_{t,i}$ are now computed as 
\begin{equation}
w_{t,i} = \max ( \weightm{t,i}^\text{fwd} \weightp{t,i} , \weightm{t,i}^\text{bwd} \weightp{t,i} ).
\label{eq:weights}
\end{equation}
We use these weights in the inference algorithm presented in \cref{sec:inference} to determine if a loop closure has occurred. 

\section{Inference}
\label{sec:inference}
To do one-dimensional magnetic field SLAM, we run a nonlinear estimation algorithm with the states and their dynamic model defined in~\cref{eq:dynModel} and, whenever a loop closure is detected, we include the measurement model defined in~\cref{eq:measModel}. If we would solve a batch, offline smoothing problem, we would first detect all landmarks~$\vl_k$, $k = 1, \hdots, K$ and could then straightforwardly solve the estimation using, \eg, Gauss-Newton optimisation, see \eg~\cite{Kok+Hol+Schon:2017,dellaertK:2006}. Our interest, however, lies in recursive estimation. Hence, at time $t = 0$, we start with time updates of an extended Kalman filter (EKF) with four states, \ie, the sensor's position, heading and the bias, without any landmarks. Since the absolute position and heading are not observable in SLAM~\cite{gustafsson:2012}, without loss of generality we initialise the position and heading to be equal to zero. We then run an EKF time update until we detect a loop closure. When detecting a loop closure, we augment our state vector with a landmark $\vl_k$ and re-run our filter with measurement updates as in \cref{eq:measModel} at time instances $t$ and $t - \ell$. Furthermore, we update the state estimates and their covariance using a Rauch-Tung-Striebel (RTS) backwards pass~\cite{rauchST:1965,sarkka:2013}. Note that the number of measurement updates is only equal to twice the number of detected loop closures. Our one-dimensional magnetic field SLAM is summarised in \cref{alg:slam}.

To detect loop closures, we compute the weights as in \cref{eq:weights} and conclude that a loop closure has occurred if $\max_i (w_{t,i}) > \gamma$. If a loop closure is detected, the time instance at which this maximum is reached is the time instance $t - \ell$ that we use in our measurement model in \cref{eq:measModel}. Since loop closures are about re-visiting locations, we do not consider potential loop closures within the most recent $N_\text{lag}$ samples. In fact, allowing loop closures to be detected in these most recent time instances would degrade our estimation results unless the user is indeed standing still. 

To increase the performance of our algorithm, we allow for discarding detected loop closures for three reasons: 
\begin{enumerate}
\item Since the computational complexity of our filter increases with additional loop closures, we only include loop closures if they are at least $N_\text{dist}$ samples from the previous loop closure. 
\item If the magnetic field is not location-dependent, \ie, there are no magnetic anomalies in the environment, the magnetic field weights in \cref{eq:magWeights} will be high at any location and loop closures can erroneously be detected. We therefore only accept a loop closure when the norm of the difference between the maximum and minimum value of $\vy_{\text{m},t-n}$ for $n = 0,1, \hdots, N_\text{lc}$ is above a threshold $\gamma_\text{mag}$. 
\item Erroneous loop closures can still occur in challenging situations in which the excitation of the magnetic field is just above the threshold $\gamma_\text{mag}$. Because of this, at the time $t$ of a detected loop closure we compute the marginal likelihood 
\begin{equation}
p(\vy_t) = \tfrac{1}{2 \pi \sqrt{|\MS_t|} } \exp \left(-\tfrac{1}{2} \vepsilon_t \T \MS^{-1}_t \vepsilon_t \right)
\end{equation}
where $\vepsilon_t$ is the difference between the predicted position and the predicted landmark location and $\MS_t$ is the residual covariance. We reject the loop closure when the marginal likelihood is smaller than a threshold $\gamma_\text{ml}$. This strategy will reject loop closures that will bend the path too significantly.
\end{enumerate}

\begin{algorithm}[t]
\caption{Magnetic field SLAM in 1D}
\label{alg:slam}
\begin{minipage}{\linewidth-14.45pt}
\begin{algorithmic}[1]
\REQUIRE Measurements $\{ \vect{y}_{\text{m},t}^\bframe, \vy_{\text{p},t}, y_{\omega,t} \}_{t = 1}^{N_T}$, parameters $\sigma_\text{p}, \sigma_\omega, \sigma_\text{lc}, \sigma_\text{m}, \gamma, N_\text{lc}, N_\text{lag}, N_\text{dist}, \gamma_\text{mag}, \gamma_\text{ml}$, and initial covariances $P_{\text{p},0}$, $P_{\psi,0}$, $P_{b_\omega,0}$, $P_{\text{l},0}$.
\ENSURE Recursively estimated state trajectories $\hat{\vx}_{0:t \mid t}$, $t = 0, \hdots, N_T$.
\STATE Set the number of landmarks $K =0$.
\STATE Initialise $\hat{\vx}_0 = \begin{pmatrix}\hat{\vp}_0\T & \hat{\psi}_0 & \hat{b}_{\omega,0} & \hat{\vl}_{1:K,0}\T\end{pmatrix}\T \in \mathbb{R}^{4+2K}$ with $\hat{\vp}_0 = \begin{pmatrix} 0 & 0 \end{pmatrix}\T$, $\hat{\psi}_0 = 0$, $\hat{b}_{\omega,0} = 0$ and $\hat{\vl}_{k,0} = \begin{pmatrix} 0 & 0 \end{pmatrix}\T, k = 1, \hdots, K$. Also create a block diagonal initial state covariance $P_0 = \text{blkdiag}\left( P_{\text{p},0}, P_{\psi,0}, P_{b_\omega,0}\right)$ and include $K$ blocks $P_{\text{l},0}$ such that $P_0 \in \mathbb{R}^{(4 + 2K) \times (4+2K)}$. \label{step:init}
\FOR{$t = 1,2, \hdots, N_T$}
\STATE Update the state estimate $\hat{\vx}_{t +1 \mid t}$ using~\cref{eq:dynModel} and update the covariance using a Kalman filter time update. 
\STATE Compute the weights~\cref{eq:weights} for $i = 1, \hdots , t - N_\text{lag}$. 
\STATE A loop closure is detected if $\max_i w_{t,i} > \gamma$ and if reasons 1) and 2) in \cref{sec:inference} are not true. 
\IF{a loop closure is detected} 
\STATE Augment the state vector with an additional landmark $\vl_k$. \label{step:loopClosureBegin}
\STATE Rerun the filter from Step~\ref{step:init} from time $\tau = 0$ until $\tau = t$. If $p(\vy_t) < \gamma_\text{ml}$, undo the loop closure. 
\STATE Run an RTS backwards smoother. \label{step:loopClosureEnd}
\ENDIF 
\ENDFOR
\end{algorithmic}
\end{minipage}
\end{algorithm}

\section{Experiments}
\label{sec:experiments}
We use a handheld iPad to collect visual-inertial odometry and magnetic field data using an in-house developed app\footnote{\url{https://github.com/asolin/ios-data-collection}} that captures the output from the Apple ARkit API. We use the accurate visual-inertial odometry to obtain information about the change in position and heading, and then disturb this by adding a bias and noise according to~\cref{eq:dynModel}, resulting in odometry measurements $\{ \vy_{\text{p},t}^\bframe, y_{\omega,t} \}_{t = 1}^{N_T}$ that can be used in \cref{alg:slam}. Unless stated otherwise, the added bias is $0.005$ rad/s, and the added noise on the position and the angular velocity has a standard deviation of $\sigma_\text{p} = 1 \cdot 10^{-2} \mathcal{I}_2$ m and $\sigma_\omega = 1 \cdot 10^{-2}$ rad/s. The magnetic field measurements are collected in a sensor frame that does not have a completely fixed inclination. Because of this, we preprocess the magnetometer measurements to obtain $\{ \vy_{\text{p},t}^\bframe \}_{t = 1}^{N_T}$ using inclination information based on ARkit. We have downsampled all data to $10$~Hz. 

\begin{figure}
	\centering
	\setlength{\figurewidth}{0.87\columnwidth}
	\setlength{\figureheight}{\columnwidth}
	\input{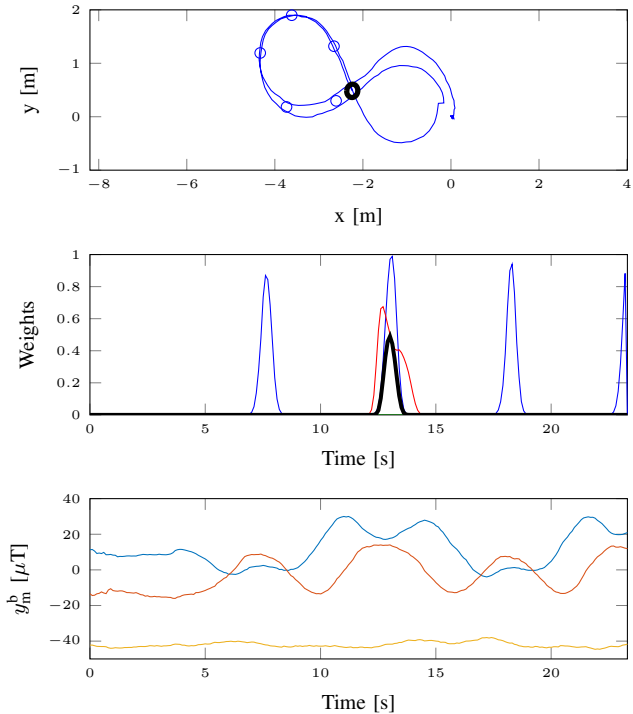}\\[-1em]
	\caption{Results on empirical proof-of-concept data with (top) the trajectory including circles indicating the locations of detected loop closures, (middle) the weights $w_{\text{m},t}^\text{fwd}$ in red, $w_{\text{p},t}$ in blue and the overall weights $w_t$ used for loop closures in black, and (bottom) the magnetic field measurements in microtesla.}
	\label{fig:eight}
\end{figure}

The values that we used for the tuning parameters of our algorithm (unless otherwise stated) are summarised in \cref{tab:tuningParams}. Choosing $N_\text{lc} = 10$, $N_\text{lag} = 50$ and $N_\text{dist} = 10$ implies that we use $1$ second of data for detecting loop closures, avoid loop closures on the most recent $5$ seconds of data, and that the loop closures need to at least be $1$ second apart. As can be seen in \cref{fig:eight}, the magnitude of the magnetic field measurements is around $45~\mu$T. Our choice of $\sigma_\text{m} = 3~\mu$T therefore implies that we assume a signal-to-noise ratio of around $15$. Furthermore, choosing $\gamma_\text{mag} = 3~\mu$T implies that we consider the excitation in the magnetic field sufficient if the norm of the difference between the maximum and minimum value of the $N_\text{lc}$ magnetometer readings is above $\sigma_\text{m}$. The weights from \cref{eq:weights} lie between $0$ and $1$. We choose a threshold of $\gamma = 0.25$ for accepting a loop closure. The small value for $\gamma_\text{ml}$ ($1 \cdot 10^{-16}$) implies that we accept almost all loop closures except when it severely bends the path. Furthermore, the small value for $\sigma_\text{lc}$ ($\sqrt{0.1}$ m) implies that we assume that when a loop closure is detected between two time instances, the positions at these time instances are very close to each other. 

To initialise our algorithm, we set initial small position and heading covariances of $P_{\text{p},0} = 1 \cdot 10^{-8} \mathcal{I}_2$ m$^2$ and $P_{\psi,0} = 1 \cdot 10^{-8}$ rad$^2$ to account for the unobservability of the SLAM problem, see \cref{sec:inference}, and $P_{b_\omega,0} = 1 \cdot 10^{-4}$ rad$^2$/s$^2$. Furthermore, we choose $P_{\text{l},0} = 1 \cdot 10^{4}$ m$^2$ to account for the large initial uncertainty of the location of the landmarks. 

\begin{table}
\caption{Values of the tuning parameters used in \cref{sec:experiments}.}
\centering
\begin{tabular}{cccccccc}\toprule
           $N_\text{lc}$ & $N_\text{lag}$ & $N_\text{dist}$ & $\sigma_\text{m}$ & $\gamma_\text{mag}$ & $\gamma$ & $\gamma_\text{ml}$ & $\sigma_\text{lc}$ \\\midrule
           $10$ & $50$ & $10$ & $3$ & $3$ & $0.25$ & $1 \cdot 10^{-16}$ & $\sqrt{0.1}$
 \\
\bottomrule
\end{tabular}
\label{tab:tuningParams}
\end{table}

\begin{figure}
	\centering
	\setlength{\figurewidth}{0.87\columnwidth}
	\setlength{\figureheight}{0.6\columnwidth}
	\pgfplotsset{axis on top,tick label style={font=\tiny},ylabel near ticks}
	\input{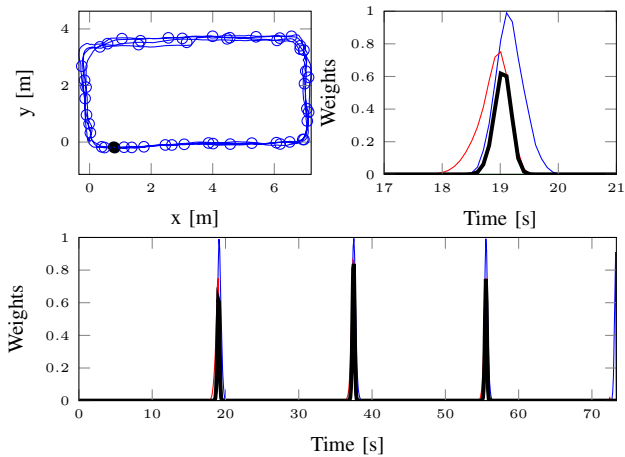}\\[-1em]
	\caption{Results on empirical proof-of-concept data with (top left) the trajectory including circles indicating the locations of detected loop closures, and (bottom and top right) the weights $w_{\text{m},t}^\text{fwd}$ in red, $w_{\text{p},t}$ in blue and the overall weights $w_t$ used for loop closures in black. These weights are shown for the entire trajectory (bottom) and zoomed in (top right).\looseness-1}
	\label{fig:square}
\end{figure}

\begin{figure*}
	\centering
	\setlength{\figurewidth}{0.5\columnwidth}
	\setlength{\figureheight}{0.5\columnwidth}
\begin{tikzpicture}

\begin{axis}[%
width=0.951\figurewidth,
height=\figureheight,
at={(0\figurewidth,0\figureheight)},
scale only axis,
xmin=0.5,
xmax=6.5,
xtick={1,2,3,4,5,6},
xticklabels={{0},{0.001},{0.005},{0.01},{0.05},{0.1}},
xlabel style={font=\color{white!15!black}},
xlabel={$\text{b}_\omega\vphantom{\sigma^2_p}$ [rad/s]},
ymode=log,
ymin=0.05,
ymax=4,
yminorticks=true,
ylabel style={font=\color{white!15!black}},
ylabel={RMSE [m]},
axis background/.style={fill=white},
ylabel style={font=\footnotesize},xlabel style={font=\footnotesize},yticklabel style={font=\tiny},xticklabel style={font=\tiny},xticklabels = {0,1E-3,5E-3,1E-2,5E-2,1E-1}
]
\addplot [color=black, dashed, forget plot]
  table[row sep=crcr]{%
1	0.133095579134699\\
1	0.163479834522367\\
};
\addplot [color=black, dashed, forget plot]
  table[row sep=crcr]{%
2	0.136638984916506\\
2	0.174043546312431\\
};
\addplot [color=black, dashed, forget plot]
  table[row sep=crcr]{%
3	0.137688408681846\\
3	0.170910900095392\\
};
\addplot [color=black, dashed, forget plot]
  table[row sep=crcr]{%
4	0.139040311738761\\
4	0.160767138011647\\
};
\addplot [color=black, dashed, forget plot]
  table[row sep=crcr]{%
5	3.91608443331758\\
5	3.92103779576941\\
};
\addplot [color=black, dashed, forget plot]
  table[row sep=crcr]{%
6	3.95415441727182\\
6	3.95528418661426\\
};
\addplot [color=black, dashed, forget plot]
  table[row sep=crcr]{%
1	0.0922552815697581\\
1	0.108602970968644\\
};
\addplot [color=black, dashed, forget plot]
  table[row sep=crcr]{%
2	0.0933788281564779\\
2	0.111114770574882\\
};
\addplot [color=black, dashed, forget plot]
  table[row sep=crcr]{%
3	0.0953604241347887\\
3	0.112178785530266\\
};
\addplot [color=black, dashed, forget plot]
  table[row sep=crcr]{%
4	0.100017138288739\\
4	0.118952962954417\\
};
\addplot [color=black, dashed, forget plot]
  table[row sep=crcr]{%
5	3.90561899285122\\
5	3.91036703935176\\
};
\addplot [color=black, dashed, forget plot]
  table[row sep=crcr]{%
6	3.95063956636027\\
6	3.95271547662173\\
};
\addplot [color=black, forget plot]
  table[row sep=crcr]{%
0.875	0.163479834522367\\
1.125	0.163479834522367\\
};
\addplot [color=black, forget plot]
  table[row sep=crcr]{%
1.875	0.174043546312431\\
2.125	0.174043546312431\\
};
\addplot [color=black, forget plot]
  table[row sep=crcr]{%
2.875	0.170910900095392\\
3.125	0.170910900095392\\
};
\addplot [color=black, forget plot]
  table[row sep=crcr]{%
3.875	0.160767138011647\\
4.125	0.160767138011647\\
};
\addplot [color=black, forget plot]
  table[row sep=crcr]{%
4.875	3.92103779576941\\
5.125	3.92103779576941\\
};
\addplot [color=black, forget plot]
  table[row sep=crcr]{%
5.875	3.95528418661426\\
6.125	3.95528418661426\\
};
\addplot [color=black, forget plot]
  table[row sep=crcr]{%
0.875	0.0922552815697581\\
1.125	0.0922552815697581\\
};
\addplot [color=black, forget plot]
  table[row sep=crcr]{%
1.875	0.0933788281564779\\
2.125	0.0933788281564779\\
};
\addplot [color=black, forget plot]
  table[row sep=crcr]{%
2.875	0.0953604241347887\\
3.125	0.0953604241347887\\
};
\addplot [color=black, forget plot]
  table[row sep=crcr]{%
3.875	0.100017138288739\\
4.125	0.100017138288739\\
};
\addplot [color=black, forget plot]
  table[row sep=crcr]{%
4.875	3.90561899285122\\
5.125	3.90561899285122\\
};
\addplot [color=black, forget plot]
  table[row sep=crcr]{%
5.875	3.95063956636027\\
6.125	3.95063956636027\\
};
\addplot [color=blue, forget plot]
  table[row sep=crcr]{%
0.75	0.108602970968644\\
0.75	0.133095579134699\\
1.25	0.133095579134699\\
1.25	0.108602970968644\\
0.75	0.108602970968644\\
};
\addplot [color=blue, forget plot]
  table[row sep=crcr]{%
1.75	0.111114770574882\\
1.75	0.136638984916506\\
2.25	0.136638984916506\\
2.25	0.111114770574882\\
1.75	0.111114770574882\\
};
\addplot [color=blue, forget plot]
  table[row sep=crcr]{%
2.75	0.112178785530266\\
2.75	0.137688408681846\\
3.25	0.137688408681846\\
3.25	0.112178785530266\\
2.75	0.112178785530266\\
};
\addplot [color=blue, forget plot]
  table[row sep=crcr]{%
3.75	0.118952962954417\\
3.75	0.139040311738761\\
4.25	0.139040311738761\\
4.25	0.118952962954417\\
3.75	0.118952962954417\\
};
\addplot [color=blue, forget plot]
  table[row sep=crcr]{%
4.75	3.91036703935176\\
4.75	3.91608443331758\\
5.25	3.91608443331758\\
5.25	3.91036703935176\\
4.75	3.91036703935176\\
};
\addplot [color=blue, forget plot]
  table[row sep=crcr]{%
5.75	3.95271547662173\\
5.75	3.95415441727182\\
6.25	3.95415441727182\\
6.25	3.95271547662173\\
5.75	3.95271547662173\\
};
\addplot [color=red, forget plot]
  table[row sep=crcr]{%
0.75	0.119149881531747\\
1.25	0.119149881531747\\
};
\addplot [color=red, forget plot]
  table[row sep=crcr]{%
1.75	0.12088163315815\\
2.25	0.12088163315815\\
};
\addplot [color=red, forget plot]
  table[row sep=crcr]{%
2.75	0.121576709341175\\
3.25	0.121576709341175\\
};
\addplot [color=red, forget plot]
  table[row sep=crcr]{%
3.75	0.130185381194206\\
4.25	0.130185381194206\\
};
\addplot [color=red, forget plot]
  table[row sep=crcr]{%
4.75	3.91337747101732\\
5.25	3.91337747101732\\
};
\addplot [color=red, forget plot]
  table[row sep=crcr]{%
5.75	3.95348300325662\\
6.25	3.95348300325662\\
};
\addplot [color=black, only marks, mark=+, mark options={solid, draw=red}, forget plot]
  table[row sep=crcr]{%
1	0.198850662198392\\
1	0.204281958842724\\
1	0.205506086809366\\
};
\addplot [color=black, only marks, mark=+, mark options={solid, draw=red}, forget plot]
  table[row sep=crcr]{%
2	0.179235797199939\\
2	0.179707997750441\\
2	0.191837059211049\\
2	0.196000301980497\\
};
\addplot [color=black, only marks, mark=+, mark options={solid, draw=red}, forget plot]
  table[row sep=crcr]{%
3	0.18380638800586\\
3	0.183876830372057\\
3	0.197214391253269\\
3	0.203374526187341\\
3	0.222275788191982\\
3	0.396426749711744\\
3	0.585349914052783\\
3	0.710406704887589\\
3	1.03458443765325\\
3	1.04653822296745\\
};
\addplot [color=black, only marks, mark=+, mark options={solid, draw=red}, forget plot]
  table[row sep=crcr]{%
4	0.0888109154790982\\
4	0.176291170175636\\
4	0.192306821019966\\
4	0.557965167293878\\
4	0.661010344687705\\
4	0.727731543839705\\
4	1.08410099802143\\
4	1.10761887691313\\
4	1.1824197032194\\
4	1.57260588289612\\
};
\addplot [color=black, only marks, mark=+, mark options={solid, draw=red}, forget plot]
  table[row sep=crcr]{%
5	3.90021703588201\\
5	3.90176365505333\\
5	3.92888103545875\\
5	3.9359008281026\\
};
\addplot [color=black, only marks, mark=+, mark options={solid, draw=red}, forget plot]
  table[row sep=crcr]{%
6	3.95052270843295\\
};
\end{axis}
\end{tikzpicture}%
\hfill
\begin{tikzpicture}

\begin{axis}[%
width=0.951\figurewidth,
height=\figureheight,
at={(0\figurewidth,0\figureheight)},
scale only axis,
unbounded coords=jump,
xmin=0.5,
xmax=7.5,
xtick={1,2,3,4,5,6,7},
xticklabels={{1e-06},{1e-05},{0.0001},{0.001},{0.005},{0.01},{0.1}},
xlabel style={font=\color{white!15!black}},
xlabel={$\sigma_\text{p}^2$ [m$^2$]},
ymode=log,
ymin=0.05,
ymax=4,
yminorticks=true,
ylabel style={font=\color{white!15!black}},
ylabel={RMSE [m]},
axis background/.style={fill=white},
ylabel style={font=\footnotesize},xlabel style={font=\footnotesize},yticklabel style={font=\tiny},xticklabel style={font=\tiny},xticklabels = {1E-6,1E-5,1E-4,1E-3,5E-2,1E-2,1E-1}
]
\addplot [color=black, dashed, forget plot]
  table[row sep=crcr]{%
1	0.122053263439857\\
1	0.144193191549516\\
};
\addplot [color=black, dashed, forget plot]
  table[row sep=crcr]{%
2	0.122784254804603\\
2	0.140171584408427\\
};
\addplot [color=black, dashed, forget plot]
  table[row sep=crcr]{%
3	0.133889414254708\\
3	0.163100469790981\\
};
\addplot [color=black, dashed, forget plot]
  table[row sep=crcr]{%
4	0.433336505007719\\
4	0.86451304983464\\
};
\addplot [color=black, dashed, forget plot]
  table[row sep=crcr]{%
5	0.981810685898917\\
5	1.88173934422183\\
};
\addplot [color=black, dashed, forget plot]
  table[row sep=crcr]{%
6	1.13306180213619\\
6	1.99747972866816\\
};
\addplot [color=black, dashed, forget plot]
  table[row sep=crcr]{%
7	2.67680985928008\\
7	3.60301120672624\\
};
\addplot [color=black, dashed, forget plot]
  table[row sep=crcr]{%
1	0.0811977340752587\\
1	0.104114504017241\\
};
\addplot [color=black, dashed, forget plot]
  table[row sep=crcr]{%
2	0.0856657890405813\\
2	0.106873146691848\\
};
\addplot [color=black, dashed, forget plot]
  table[row sep=crcr]{%
3	0.0929483303323361\\
3	0.114236449869277\\
};
\addplot [color=black, dashed, forget plot]
  table[row sep=crcr]{%
4	0.107628556090401\\
4	0.14476661695457\\
};
\addplot [color=black, dashed, forget plot]
  table[row sep=crcr]{%
5	0.150815127789801\\
5	0.355083659474358\\
};
\addplot [color=black, dashed, forget plot]
  table[row sep=crcr]{%
6	0.177427119273164\\
6	0.554062238367916\\
};
\addplot [color=black, dashed, forget plot]
  table[row sep=crcr]{%
7	1.11502666060424\\
7	1.95491745985889\\
};
\addplot [color=black, forget plot]
  table[row sep=crcr]{%
0.875	0.144193191549516\\
1.125	0.144193191549516\\
};
\addplot [color=black, forget plot]
  table[row sep=crcr]{%
1.875	0.140171584408427\\
2.125	0.140171584408427\\
};
\addplot [color=black, forget plot]
  table[row sep=crcr]{%
2.875	0.163100469790981\\
3.125	0.163100469790981\\
};
\addplot [color=black, forget plot]
  table[row sep=crcr]{%
3.875	0.86451304983464\\
4.125	0.86451304983464\\
};
\addplot [color=black, forget plot]
  table[row sep=crcr]{%
4.875	1.88173934422183\\
5.125	1.88173934422183\\
};
\addplot [color=black, forget plot]
  table[row sep=crcr]{%
5.875	1.99747972866816\\
6.125	1.99747972866816\\
};
\addplot [color=black, forget plot]
  table[row sep=crcr]{%
6.875	3.60301120672624\\
7.125	3.60301120672624\\
};
\addplot [color=black, forget plot]
  table[row sep=crcr]{%
0.875	0.0811977340752587\\
1.125	0.0811977340752587\\
};
\addplot [color=black, forget plot]
  table[row sep=crcr]{%
1.875	0.0856657890405813\\
2.125	0.0856657890405813\\
};
\addplot [color=black, forget plot]
  table[row sep=crcr]{%
2.875	0.0929483303323361\\
3.125	0.0929483303323361\\
};
\addplot [color=black, forget plot]
  table[row sep=crcr]{%
3.875	0.107628556090401\\
4.125	0.107628556090401\\
};
\addplot [color=black, forget plot]
  table[row sep=crcr]{%
4.875	0.150815127789801\\
5.125	0.150815127789801\\
};
\addplot [color=black, forget plot]
  table[row sep=crcr]{%
5.875	0.177427119273164\\
6.125	0.177427119273164\\
};
\addplot [color=black, forget plot]
  table[row sep=crcr]{%
6.875	1.11502666060424\\
7.125	1.11502666060424\\
};
\addplot [color=blue, forget plot]
  table[row sep=crcr]{%
0.75	0.104114504017241\\
0.75	0.122053263439857\\
1.25	0.122053263439857\\
1.25	0.104114504017241\\
0.75	0.104114504017241\\
};
\addplot [color=blue, forget plot]
  table[row sep=crcr]{%
1.75	0.106873146691848\\
1.75	0.122784254804603\\
2.25	0.122784254804603\\
2.25	0.106873146691848\\
1.75	0.106873146691848\\
};
\addplot [color=blue, forget plot]
  table[row sep=crcr]{%
2.75	0.114236449869277\\
2.75	0.133889414254708\\
3.25	0.133889414254708\\
3.25	0.114236449869277\\
2.75	0.114236449869277\\
};
\addplot [color=blue, forget plot]
  table[row sep=crcr]{%
3.75	0.14476661695457\\
3.75	0.433336505007719\\
4.25	0.433336505007719\\
4.25	0.14476661695457\\
3.75	0.14476661695457\\
};
\addplot [color=blue, forget plot]
  table[row sep=crcr]{%
4.75	0.355083659474358\\
4.75	0.981810685898917\\
5.25	0.981810685898917\\
5.25	0.355083659474358\\
4.75	0.355083659474358\\
};
\addplot [color=blue, forget plot]
  table[row sep=crcr]{%
5.75	0.554062238367916\\
5.75	1.13306180213619\\
6.25	1.13306180213619\\
6.25	0.554062238367916\\
5.75	0.554062238367916\\
};
\addplot [color=blue, forget plot]
  table[row sep=crcr]{%
6.75	1.95491745985889\\
6.75	2.67680985928008\\
7.25	2.67680985928008\\
7.25	1.95491745985889\\
6.75	1.95491745985889\\
};
\addplot [color=red, forget plot]
  table[row sep=crcr]{%
0.75	0.109583061060836\\
1.25	0.109583061060836\\
};
\addplot [color=red, forget plot]
  table[row sep=crcr]{%
1.75	0.112551774362105\\
2.25	0.112551774362105\\
};
\addplot [color=red, forget plot]
  table[row sep=crcr]{%
2.75	0.122625574676425\\
3.25	0.122625574676425\\
};
\addplot [color=red, forget plot]
  table[row sep=crcr]{%
3.75	0.180751036416441\\
4.25	0.180751036416441\\
};
\addplot [color=red, forget plot]
  table[row sep=crcr]{%
4.75	0.576279971064591\\
5.25	0.576279971064591\\
};
\addplot [color=red, forget plot]
  table[row sep=crcr]{%
5.75	0.802316214206916\\
6.25	0.802316214206916\\
};
\addplot [color=red, forget plot]
  table[row sep=crcr]{%
6.75	2.36335430924889\\
7.25	2.36335430924889\\
};
\addplot [color=black, only marks, mark=+, mark options={solid, draw=red}, forget plot]
  table[row sep=crcr]{%
1	0.14897879188231\\
1	0.155622668306675\\
1	0.176208725344138\\
1	0.177977444781767\\
1	0.182200746864623\\
1	0.282884992774943\\
1	0.450918986260982\\
1	0.679813637551505\\
1	1.00138409476064\\
};
\addplot [color=black, only marks, mark=+, mark options={solid, draw=red}, forget plot]
  table[row sep=crcr]{%
2	0.159345947215667\\
2	0.160615461444236\\
2	0.161595356548798\\
2	0.163789281123242\\
2	0.191955274034026\\
2	0.245477270929552\\
2	0.968853967492384\\
};
\addplot [color=black, only marks, mark=+, mark options={solid, draw=red}, forget plot]
  table[row sep=crcr]{%
3	0.166956316319363\\
3	0.17868887448026\\
3	0.197523037882824\\
3	0.218142109867042\\
3	0.220579626695205\\
3	0.414989357882773\\
3	0.587176430841159\\
3	0.82181831535798\\
3	0.995495044239216\\
};
\addplot [color=black, only marks, mark=+, mark options={solid, draw=red}, forget plot]
  table[row sep=crcr]{%
4	1.05703841426093\\
4	1.13035696710142\\
4	1.15548708409393\\
4	1.206230601974\\
4	1.23018028285512\\
4	1.23700088462671\\
4	1.46166200841261\\
4	1.53440981375907\\
4	1.54094887827564\\
4	1.63541521519979\\
4	1.72970910137245\\
4	1.78699662850728\\
};
\addplot [color=black, only marks, mark=+, mark options={solid, draw=red}, forget plot]
  table[row sep=crcr]{%
5	1.94498110777366\\
5	1.96337377251115\\
5	2.00734445717178\\
5	2.03875268692372\\
5	2.09168053377471\\
5	2.26459161105149\\
};
\addplot [color=black, only marks, mark=+, mark options={solid, draw=red}, forget plot]
  table[row sep=crcr]{%
6	2.35950534724143\\
6	2.55002961439424\\
};
\addplot [color=black, only marks, mark=+, mark options={solid, draw=red}, forget plot]
  table[row sep=crcr]{%
nan	nan\\
};
\end{axis}
\end{tikzpicture}%
\hfill
\begin{tikzpicture}

\begin{axis}[%
width=0.951\figurewidth,
height=\figureheight,
at={(0\figurewidth,0\figureheight)},
scale only axis,
unbounded coords=jump,
xmin=0.5,
xmax=7.5,
xtick={1,2,3,4,5,6,7},
xticklabels={{1e-06},{1e-05},{0.0001},{0.001},{0.005},{0.01},{0.1}},
xlabel style={font=\color{white!15!black}},
xlabel={$\sigma_\omega^2$ [rad$^2$/s$^2$]},
ymode=log,
ymin=0.05,
ymax=4,
yminorticks=true,
ylabel style={font=\color{white!15!black}},
ylabel={RMSE [m]},
axis background/.style={fill=white},
ylabel style={font=\footnotesize},xlabel style={font=\footnotesize},yticklabel style={font=\tiny},xticklabel style={font=\tiny},xticklabels = {1E-6,1E-5,1E-4,1E-3,5E-2,1E-2,1E-1}
]
\addplot [color=black, dashed, forget plot]
  table[row sep=crcr]{%
1	0.081179151462823\\
1	0.115123263709147\\
};
\addplot [color=black, dashed, forget plot]
  table[row sep=crcr]{%
2	0.0937863198963207\\
2	0.109585186767462\\
};
\addplot [color=black, dashed, forget plot]
  table[row sep=crcr]{%
3	0.136794081384915\\
3	0.17092368601022\\
};
\addplot [color=black, dashed, forget plot]
  table[row sep=crcr]{%
4	0.711879448755596\\
4	1.39998001357512\\
};
\addplot [color=black, dashed, forget plot]
  table[row sep=crcr]{%
5	1.34029494940599\\
5	1.65849171554658\\
};
\addplot [color=black, dashed, forget plot]
  table[row sep=crcr]{%
6	1.71369644885178\\
6	2.31274495566689\\
};
\addplot [color=black, dashed, forget plot]
  table[row sep=crcr]{%
7	3.29285419226036\\
7	3.70070035558849\\
};
\addplot [color=black, dashed, forget plot]
  table[row sep=crcr]{%
1	0.0413617421895104\\
1	0.0545420710451889\\
};
\addplot [color=black, dashed, forget plot]
  table[row sep=crcr]{%
2	0.0501081705201571\\
2	0.0705257586894273\\
};
\addplot [color=black, dashed, forget plot]
  table[row sep=crcr]{%
3	0.0829437693484042\\
3	0.113442285094833\\
};
\addplot [color=black, dashed, forget plot]
  table[row sep=crcr]{%
4	0.177067892798546\\
4	0.243029419049412\\
};
\addplot [color=black, dashed, forget plot]
  table[row sep=crcr]{%
5	0.856276727549655\\
5	1.08503193974479\\
};
\addplot [color=black, dashed, forget plot]
  table[row sep=crcr]{%
6	0.855506499061093\\
6	1.30124384417863\\
};
\addplot [color=black, dashed, forget plot]
  table[row sep=crcr]{%
7	2.15943284574199\\
7	2.67602167007965\\
};
\addplot [color=black, forget plot]
  table[row sep=crcr]{%
0.875	0.115123263709147\\
1.125	0.115123263709147\\
};
\addplot [color=black, forget plot]
  table[row sep=crcr]{%
1.875	0.109585186767462\\
2.125	0.109585186767462\\
};
\addplot [color=black, forget plot]
  table[row sep=crcr]{%
2.875	0.17092368601022\\
3.125	0.17092368601022\\
};
\addplot [color=black, forget plot]
  table[row sep=crcr]{%
3.875	1.39998001357512\\
4.125	1.39998001357512\\
};
\addplot [color=black, forget plot]
  table[row sep=crcr]{%
4.875	1.65849171554658\\
5.125	1.65849171554658\\
};
\addplot [color=black, forget plot]
  table[row sep=crcr]{%
5.875	2.31274495566689\\
6.125	2.31274495566689\\
};
\addplot [color=black, forget plot]
  table[row sep=crcr]{%
6.875	3.70070035558849\\
7.125	3.70070035558849\\
};
\addplot [color=black, forget plot]
  table[row sep=crcr]{%
0.875	0.0413617421895104\\
1.125	0.0413617421895104\\
};
\addplot [color=black, forget plot]
  table[row sep=crcr]{%
1.875	0.0501081705201571\\
2.125	0.0501081705201571\\
};
\addplot [color=black, forget plot]
  table[row sep=crcr]{%
2.875	0.0829437693484042\\
3.125	0.0829437693484042\\
};
\addplot [color=black, forget plot]
  table[row sep=crcr]{%
3.875	0.177067892798546\\
4.125	0.177067892798546\\
};
\addplot [color=black, forget plot]
  table[row sep=crcr]{%
4.875	0.856276727549655\\
5.125	0.856276727549655\\
};
\addplot [color=black, forget plot]
  table[row sep=crcr]{%
5.875	0.855506499061093\\
6.125	0.855506499061093\\
};
\addplot [color=black, forget plot]
  table[row sep=crcr]{%
6.875	2.15943284574199\\
7.125	2.15943284574199\\
};
\addplot [color=blue, forget plot]
  table[row sep=crcr]{%
0.75	0.0545420710451889\\
0.75	0.081179151462823\\
1.25	0.081179151462823\\
1.25	0.0545420710451889\\
0.75	0.0545420710451889\\
};
\addplot [color=blue, forget plot]
  table[row sep=crcr]{%
1.75	0.0705257586894273\\
1.75	0.0937863198963207\\
2.25	0.0937863198963207\\
2.25	0.0705257586894273\\
1.75	0.0705257586894273\\
};
\addplot [color=blue, forget plot]
  table[row sep=crcr]{%
2.75	0.113442285094833\\
2.75	0.136794081384915\\
3.25	0.136794081384915\\
3.25	0.113442285094833\\
2.75	0.113442285094833\\
};
\addplot [color=blue, forget plot]
  table[row sep=crcr]{%
3.75	0.243029419049412\\
3.75	0.711879448755596\\
4.25	0.711879448755596\\
4.25	0.243029419049412\\
3.75	0.243029419049412\\
};
\addplot [color=blue, forget plot]
  table[row sep=crcr]{%
4.75	1.08503193974479\\
4.75	1.34029494940599\\
5.25	1.34029494940599\\
5.25	1.08503193974479\\
4.75	1.08503193974479\\
};
\addplot [color=blue, forget plot]
  table[row sep=crcr]{%
5.75	1.30124384417863\\
5.75	1.71369644885178\\
6.25	1.71369644885178\\
6.25	1.30124384417863\\
5.75	1.30124384417863\\
};
\addplot [color=blue, forget plot]
  table[row sep=crcr]{%
6.75	2.67602167007965\\
6.75	3.29285419226036\\
7.25	3.29285419226036\\
7.25	2.67602167007965\\
6.75	2.67602167007965\\
};
\addplot [color=red, forget plot]
  table[row sep=crcr]{%
0.75	0.0669561879677932\\
1.25	0.0669561879677932\\
};
\addplot [color=red, forget plot]
  table[row sep=crcr]{%
1.75	0.0815329626731248\\
2.25	0.0815329626731248\\
};
\addplot [color=red, forget plot]
  table[row sep=crcr]{%
2.75	0.12376887865041\\
3.25	0.12376887865041\\
};
\addplot [color=red, forget plot]
  table[row sep=crcr]{%
3.75	0.404789043740633\\
4.25	0.404789043740633\\
};
\addplot [color=red, forget plot]
  table[row sep=crcr]{%
4.75	1.19263849477828\\
5.25	1.19263849477828\\
};
\addplot [color=red, forget plot]
  table[row sep=crcr]{%
5.75	1.47197646057124\\
6.25	1.47197646057124\\
};
\addplot [color=red, forget plot]
  table[row sep=crcr]{%
6.75	3.01507602709907\\
7.25	3.01507602709907\\
};
\addplot [color=black, only marks, mark=+, mark options={solid, draw=red}, forget plot]
  table[row sep=crcr]{%
1	0.129241668067934\\
};
\addplot [color=black, only marks, mark=+, mark options={solid, draw=red}, forget plot]
  table[row sep=crcr]{%
2	0.138942470695266\\
2	0.146881158084742\\
2	0.177292650185563\\
2	0.208298123293092\\
};
\addplot [color=black, only marks, mark=+, mark options={solid, draw=red}, forget plot]
  table[row sep=crcr]{%
3	0.178535760509972\\
3	0.178604869531659\\
3	0.24866064594887\\
3	0.385718816867416\\
3	0.574276013531956\\
3	0.706739092460702\\
3	0.834747539057092\\
3	1.17877437055809\\
};
\addplot [color=black, only marks, mark=+, mark options={solid, draw=red}, forget plot]
  table[row sep=crcr]{%
4	1.42094308680168\\
4	1.51617837863161\\
4	1.52460120370554\\
4	1.52747462656097\\
4	1.5848454791445\\
4	1.60973663919165\\
4	1.70465620578199\\
4	1.73941266884989\\
4	1.77367995355807\\
4	1.82055825919499\\
};
\addplot [color=black, only marks, mark=+, mark options={solid, draw=red}, forget plot]
  table[row sep=crcr]{%
5	1.75322354933789\\
5	1.84524608077501\\
5	1.90241909960061\\
};
\addplot [color=black, only marks, mark=+, mark options={solid, draw=red}, forget plot]
  table[row sep=crcr]{%
nan	nan\\
};
\addplot [color=black, only marks, mark=+, mark options={solid, draw=red}, forget plot]
  table[row sep=crcr]{%
nan	nan\\
};
\end{axis}
\end{tikzpicture}%
	\caption{Position RMSEs of 100 Monte Carlo simulations for different odometry accuracies: Varying gyroscope bias $b_\omega$ (left), varying position noise variance $\sigma^2_\text{p}$ (middle) and varying angular velocity noise variance $\sigma^2_\omega$ (right).}
	\label{fig:mc}
\end{figure*}
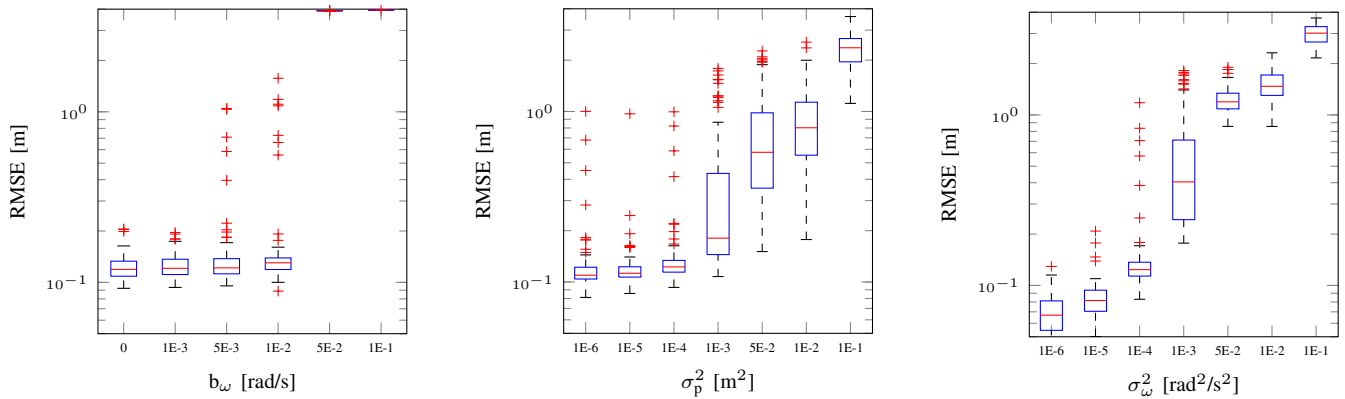

\subsection{Empirical proof-of-concept data}
To illustrate the workings of our algorithm we use two data sets in which we walk a well-defined trajectory. Firstly, we collect data while walking in a square trajectory as displayed on the left top of \cref{fig:square}. The trajectory is depicted as a blue line, the blue circles visualise the locations at which a loop closure is performed, \ie, when Lines~\ref{step:loopClosureBegin}--\ref{step:loopClosureEnd} of \cref{alg:slam} are executed. \cref{fig:square} visualises the trajectory after almost four laps. The current location is indicated by the black circle, where the magnitude of the circle represents the uncertainty of the position estimate obtained from \cref{alg:slam}. The weights computed for loop closure at the current time instance are shown in the lower plot of \cref{fig:square}. We also zoom in on a specific time period in the right upper plot. The red line represents the weights based on the magnetic field measurements as computed according to~\cref{eq:magWeights} in the forward direction; the weights in the backward direction are zero. The blue line represents the weights computed according to~\cref{eq:posWeights}. The black line depicts the overall weight computed according to~\cref{eq:weights}. It can be seen that there are indeed three time instances, corresponding to the previous three times when the same location was visited, where both the magnetic field and the location are similar and hence the overall weight is large. 

We also illustrate the workings of our algorithm for a second data set in \cref{fig:eight}. As illustrated by the trajectory of this proof-of-concept data, it is not essential for \cref{alg:slam} to walk in straight lines. Apart from the trajectory (top) and the weights (middle), we also depict the magnetic field measurements. Even though the user is at the center-point of the trajectory (black circle) and the weights~\eqref{eq:posWeights} representing the position likelihood can be seen to detect the three distinct time instances when the location has been visited, the magnetic field weights from \cref{eq:magWeights} can be seen to be able to distinguish between the different locations. 

\subsection{Small-scale experiment}
We also use our algorithm for data with a less well-defined walking pattern, as visualised in \cref{fig:library}. The collected visual-inertial trajectory can be found in \cref{fig:library} (top left). As can be seen (middle top, red), the odometry that we simulate drifts over time. In blue (top right), our estimation results can be found as well as the locations of the loop closures. We include a plot of the magnetic field measurements twice in the figure in the bottom plot. The grey lines indicate the locations at which the loop closure detection algorithm detected that the magnetic field was similar and hence a loop closure could be performed. Note that for this experiment we set $\gamma_\text{ml} = 0.1$ to avoid erroneous loop closures.

The blue and green trajectories in \cref{fig:library} can be seen to be visually very similar. Collecting ground truth data for trajectories like this is challenging as optical motion capture systems, which are typically used for ground truth due to their high accuracy, constrain the motion to much smaller areas than what we consider in this experiment. Although the visual-inertial trajectory is known to have inherent estimation errors of which the exact magnitude is unknown, visually it can still be seen that the trajectory output by ARKit and our estimated trajectory look similar. In fact, the RMS of the difference between the two position estimates is around 12 cm, which is in line with previously reported localisation accuracy of the magnetic field in the order of decimeters~\cite{Li+Gallagher+Dempster+Rizos:2012}. Note that for the comparison we first use the Procrustes algorithm to align the two trajectories more optimally, which is common within SLAM as the absolute position and heading are unobservable. To illustrate how this RMS depends on the quality of the odometry, we run 100 Monte Carlo simulations for different values of the bias $b_\omega$ and the position and angular velocity noise variance $\sigma_p^2$ and $\sigma_\omega^2$. The results are shown in \cref{fig:mc}, where it can be seen that our algorithm results in accurate position estimates when the bias and noise variances that are moderate, but for larger values our algorithm suffers from erroneous loop closure detections which can result in large estimation errors. 

\begin{figure}
\centering
\setlength{\figurewidth}{0.87\columnwidth}
\setlength{\figureheight}{0.8\columnwidth}
\pgfplotsset{axis on top}
\input{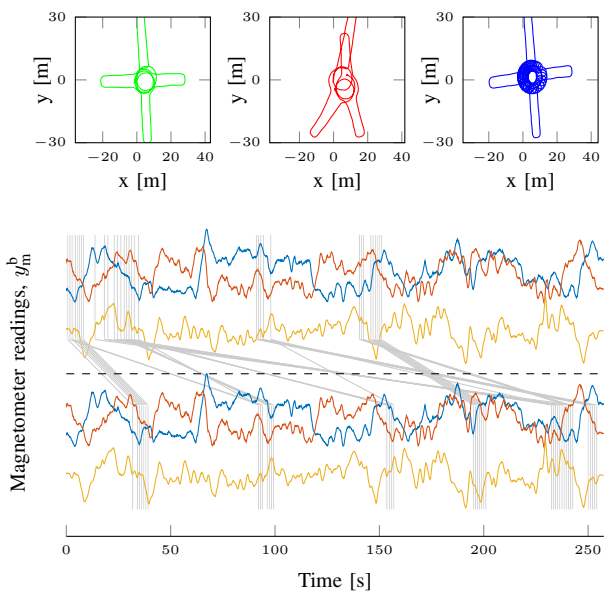}\\[-1em]
\caption{Illustration of our method with top left (green) ground truth, top middle (red) odometry, top right (blue) estimation results. In the bottom, we twice visualise the magnetic field measurements over time, above and below the black dashed line. In grey we indicate the locations where the magnetic field is detected to be similar, resulting in loop closures in \cref{alg:slam}.}
\label{fig:kamppi}
\end{figure}

\subsection{Large-scale magnetic field SLAM}
Finally, we use \cref{alg:slam} on a large-scale data set collected in the shopping mall Kamppi in Helsinki, Finland, in which we collect a path of around $350$ meters traversed during 250 seconds. For a first inefficient implementation in Matlab, it takes 32 seconds to run \cref{alg:slam} on this data set on a MacBook Pro, 16GB Apple M1. The memory requirement for our algorithm is dominated by the memory required to save a square matrix of the size of the number of states at every time instance for the RTS backwards smoother. The memory requirement for storing the past magnetometer measurements is negligible with respect to this. The estimation results can be found in \cref{fig:kamppi}. To obtain these results we set the process noise covariance in our filter to 10 times the covariance we used to simulate the odometry, to account for unmodelled errors in the data. 

The data set only allows for loop closures at the centre, as all other paths are only traversed once. This can be seen to be sufficient to remove the drift from the odometry. The magnetic field data can be seen to be sufficiently exciting in the shopping mall and also sufficiently constant over time to do the loop closures also in this real-world scenario. 

\section{Discussion and conclusions}
We have presented a method for one-dimensional magnetic field SLAM in which we detect magnetic landmarks and use these in an extended Kalman smoother to correct drift in odometry paths. Our method is computationally efficient and we have shown that it results in accurate position estimates in both large and small scale experiments with data collected using a handheld iPad. In future work it is possible to explore different ways to solve the estimation problem, \eg\ Graph SLAM~\cite{Thrun+Montemerlo:2006}. 

Our work should be seen as a proof-of-concept of how the magnetic field can be used to obtain accurate position estimates for pedestrians walking through corridors if reasonably accurate odometry information is available. Because of this, we artificially generated this odometry by perturbing paths obtained from visual-inertial odometry. An important other direction for future work is to explore using a different source of odometry information, \eg\ from pedestrian dead-reckoning algorithms instead~\cite{Wu+Zhu+Du+Tang:2019}. It would then also be interesting to study using our algorithm in different carrying modes, rather than only assuming that the user does not rotate the phone with respect to the body~\cite{tian2015multi}. Another direction of future research is to study the tuning of the parameters more in-depth. For instance, we had to choose a larger value for $\gamma_\text{ml}$ for the small-scale experiment than in the other experiments to avoid erroneous loop closures. It would be interesting to study if the value for this parameter can be chosen based on the expected number of loop closures in the data set or on the specific trajectory. 

One major limitation of our work is that, as any SLAM algorithm that uses explicit loop closures, the performance of our algorithm severely degrades when erroneous loop closures are included. This could be addressed using \eg\ a filter bank~\cite{blackmanP:1999}, in which we estimate the states assuming both that a loop closure occurred and that it did not occur, allowing for retrospectively determining if the loop closure was erroneous. Another way of addressing this limitation is by more advanced magnetic landmark detection methods. In fact, our heuristic landmark detection method can be replaced with any other, more sophisticated, method. For instance, in this work we implicitly assume a fairly constant walking speed. Although there is some natural variation in the walking speed in our data sets, an interesting direction of future work to make our method more robust is to extend it by incorporating ideas from~\cite{Montoliu+Torres-Sospedra+Belmonte:2016,subbuGD:2011} to allow for varying walking speeds.

Another interesting direction of future work would be to extend our method to 2D or 3D localisation. In these cases it will no longer be possible to use the landmark detection method from \cref{sec:detectionLandmarks}. However, after replacing the landmark detection algorithm it is straightforward to adapt \cref{alg:slam} to 2D / 3D localisation. The main challenge is that the landmarks will become much more difficult to detect as the magnetic field will be less unique. It would be interesting to explore in which scenarios 2D / 3D localisation is nevertheless a possibility, potentially by not only relying on online loop closure detections but by also incorporating information from a database of previously collected magnetic landmarks.\looseness-1 

\bibliographystyle{IEEEtran}

\begin{thebibliography}{10}
\providecommand{\url}[1]{#1}
\csname url@samestyle\endcsname
\providecommand{\newblock}{\relax}
\providecommand{\bibinfo}[2]{#2}
\providecommand{\BIBentrySTDinterwordspacing}{\spaceskip=0pt\relax}
\providecommand{\BIBentryALTinterwordstretchfactor}{4}
\providecommand{\BIBentryALTinterwordspacing}{\spaceskip=\fontdimen2\font plus
\BIBentryALTinterwordstretchfactor\fontdimen3\font minus
  \fontdimen4\font\relax}
\providecommand{\BIBforeignlanguage}[2]{{%
\expandafter\ifx\csname l@#1\endcsname\relax
\typeout{** WARNING: IEEEtran.bst: No hyphenation pattern has been}%
\typeout{** loaded for the language `#1'. Using the pattern for}%
\typeout{** the default language instead.}%
\else
\language=\csname l@#1\endcsname
\fi
#2}}
\providecommand{\BIBdecl}{\relax}
\BIBdecl

\bibitem{Haverinen+Kemppainen:2009}
J.~Haverinen and A.~Kemppainen, ``Global indoor self-localization based on the
  ambient magnetic field,'' \emph{Robotics and Autonomous Systems}, vol.~57,
  no.~10, pp. 1028--1035, 2009.

\bibitem{Li+Gallagher+Dempster+Rizos:2012}
B.~Li, T.~Gallagher, A.~G. Dempster, and C.~Rizos, ``How feasible is the use of
  magnetic field alone for indoor positioning?'' in \emph{Proceedings of the
  International Conference on Indoor Positioning and Indoor Navigation (IPIN)},
  2012, pp. 1--9.

\bibitem{Angermann+Frassl+Doniec+Julian+Robertson:2012}
M.~Angermann, M.~Frassl, M.~Doniec, B.~J. Julian, and P.~Robertson,
  ``Characterization of the indoor magnetic field for applications in
  localization and mapping,'' in \emph{Proceedings of the International
  Conference on Indoor Positioning and Indoor Navigation (IPIN)}, 2012, pp.
  1--9.

\bibitem{LeGrand+Thrun:2012}
E.~Le~Grand and S.~Thrun, ``3-axis magnetic field mapping and fusion for indoor
  localization,'' in \emph{Proceedings of the IEEE Conference on Multisensor
  Fusion and Integration for Intelligent Systems (MFI)}, 2012, pp. 358--364.

\bibitem{Solin+Sarkka+Kannala+Rahtu:2016}
A.~Solin, S.~S\"arkk\"a, J.~Kannala, and E.~Rahtu, ``Terrain navigation in the
  magnetic landscape: {P}article filtering for indoor positioning,'' in
  \emph{Proceedings of the European Navigation Conference}, 2016.

\bibitem{Hanley+Faustino+Zelman+Degenhardt+Bretl:2017}
D.~Hanley, A.~B. Faustino, S.~D. Zelman, D.~A. Degenhardt, and T.~Bretl,
  ``Mag{PIE}: {A} dataset for indoor positioning with magnetic anomalies,'' in
  \emph{Proceedings of the International Conference on Indoor Positioning and
  Indoor Navigation (IPIN)}, 2017, pp. 1--8.

\bibitem{Gao+Harle:2017}
C.~Gao and R.~Harle, ``{MSGD}: {S}calable back-end for indoor magnetic
  field-based {GraphSLAM},'' in \emph{Proceedings of the IEEE International
  Conference on Robotics and Automation (ICRA)}, 2017, pp. 3855--3862.

\bibitem{Torres+Rambla+Montoliu+Belmonte+Huerta:2015}
J.~Torres-Sospedra, D.~Rambla, R.~Montoliu, O.~Belmonte, and J.~Huerta,
  ``{UJIIndoorLoc-Mag}: {A} new database for magnetic field-based localization
  problems,'' in \emph{Proceedings of the International Conference on Indoor
  Positioning and Indoor Navigation (IPIN)}, 2015, pp. 1--10.

\bibitem{Kok+Solin:2018}
M.~Kok and A.~Solin, ``Scalable magnetic field {SLAM} in 3{D} using {G}aussian
  process maps,'' in \emph{Proceedings of the 20th International Conference on
  Information Fusion}, Cambridge, UK, July 2018.

\bibitem{visetHK:2022}
F.~M. Viset, R.~Helmons, and M.~Kok, ``An extended {K}alman filter for magnetic
  field {SLAM} using {G}aussian process regression,'' \emph{MDPI Sensors},
  vol.~22, no.~8, p. 2833, 2022.

\bibitem{ouyangA:2022}
G.~Ouyang and K.~Abed-Meraim, ``A survey of magnetic-field-based indoor
  localization,'' \emph{MDPI Electronics}, vol.~11, no.~6, p. 864, 2022.

\bibitem{kok2024rao}
M.~Kok, A.~Solin, and T.~B. Sch{\"o}n, ``Rao-{B}lackwellized particle smoothing
  for simultaneous localization and mapping,'' \emph{Data-Centric Engineering},
  vol.~5, p. e15, 2024.

\bibitem{osmanVK:2022}
M.~Osman, F.~Viset, and M.~Kok, ``Indoor {SLAM} using a foot-mounted {IMU} and
  the local magnetic field,'' in \emph{Proceedings of the 25th International
  Conference on Information Fusion}, Link\"oping, Sweden, Jul. 2022, pp. 1--7.

\bibitem{Rasmussen+Williams:2006}
C.~E. Rasmussen and C.~K.~I. Williams, \emph{Gaussian Processes for Machine
  Learning}.\hskip 1em plus 0.5em minus 0.4em\relax MIT Press, 2006.

\bibitem{Siebler+Lehner+Sand+Hanebeck:2024}
B.~Siebler, A.~Lehner, S.~Sand, and U.~D. Hanebeck, ``Magnetic field mapping of
  railway lines with graph {SLAM},'' in \emph{Proceedings of the 20th
  International Conference on Information Fusion}, Venice, Italy, July 2024.

\bibitem{suksakulchaiTWK:2000}
S.~Suksakulchai, S.~Thongchai, D.~M. Wilkes, and K.~Kawamura, ``Mobile robot
  localization using an electronic compass for corridor environment,'' in
  \emph{Proceedings of the IEEE International Conference on Systems, Man, and
  Cybernetics (SMC)}, vol.~5, Nashville, USA, October 2000, pp. 3354--3359.

\bibitem{subbuGD:2011}
K.~P. Subbu, B.~Gozick, and R.~Dantu, ``Indoor localization through dynamic
  time warping,'' in \emph{Proceedings of the IEEE International Conference on
  Systems, Man, and Cybernetics}, Anchorage, AK, USA, Oct. 2011, pp.
  1639--1644.

\bibitem{Montoliu+Torres-Sospedra+Belmonte:2016}
R.~Montoliu, J.~Torres-Sospedra, and O.~Belmonte, ``Magnetic field based indoor
  positioning using the bag of words paradigm,'' in \emph{Proceedings of the
  International Conference on Indoor Positioning and Indoor Navigation (IPIN)},
  Alcala de Henares, Spain, Oct. 2016.

\bibitem{leeAH:2018}
N.~Lee, S.~Ahn, and D.~Han, ``{AMID: A}ccurate magnetic indoor localization
  using deep learning,'' \emph{MDPI Sensors}, vol.~18, no.~5, p. 1598, 2018.

\bibitem{carrilloMUS:2015}
D.~Carrillo, V.~Moreno, B.~{\'U}beda, and A.~F. Skarmeta, ``Magic{F}inger: 3{D}
  magnetic fingerprints for indoor location,'' \emph{MDPI Sensors}, vol.~15,
  no.~7, pp. 17\,168--17\,194, 2015.

\bibitem{wangJDCWHA:2024}
Q.~Wang, J.~Jia, Y.~Deng, J.~Chen, X.~Wang, M.~Huang, and A.~H. Aghvami,
  ``{DarLoc: D}eep learning and data-feature augmentation based robust magnetic
  indoor localization,'' \emph{Expert Systems with Applications}, vol. 244, p.
  122921, 2024.

\bibitem{Kok+Hol+Schon:2017}
M.~Kok, J.~D. Hol, and T.~B. Sch\"on, ``Using inertial sensors for position and
  orientation estimation,'' \emph{Foundations and Trends on Signal Processing},
  vol.~11, no. 1--2, pp. 1--153, 2017.

\bibitem{dellaertK:2006}
F.~Dellaert and M.~Kaess, ``Square root {SAM}: {S}imultaneous localization and
  mapping via square root information smoothing,'' \emph{The International
  Journal of Robotics Research}, vol.~25, no.~12, pp. 1181--1203, 2006.

\bibitem{gustafsson:2012}
F.~Gustafsson, \emph{Statistical Sensor Fusion}.\hskip 1em plus 0.5em minus
  0.4em\relax Studentlitteratur, 2012.

\bibitem{rauchST:1965}
H.~E. Rauch, C.~T. Striebel, and F.~Tung, ``Maximum likelihood estimates of
  linear dynamic systems,'' \emph{AIAA journal}, vol.~3, no.~8, pp. 1445--1450,
  1965.

\bibitem{sarkka:2013}
S.~S{\"a}rkk{\"a}, \emph{Bayesian Filtering and Smoothing}.\hskip 1em plus
  0.5em minus 0.4em\relax Cambridge University Press, 2013.

\bibitem{Thrun+Montemerlo:2006}
S.~Thrun and M.~Montemerlo, ``The graph {SLAM} algorithm with applications to
  large-scale mapping of urban structures,'' \emph{The International Journal of
  Robotics Research}, vol.~25, no. 5-6, pp. 403--429, 2006.

\bibitem{Wu+Zhu+Du+Tang:2019}
Y.~Wu, H.-B. Zhu, Q.-X. Du, and S.-M. Tang, ``A survey of the research status
  of pedestrian dead reckoning systems based on inertial sensors,''
  \emph{International Journal of Automation and Computing}, vol.~16, pp.
  65--83, 2019.

\bibitem{tian2015multi}
Q.~Tian, Z.~Salcic, I.~Kevin, K.~Wang, and Y.~Pan, ``A multi-mode dead
  reckoning system for pedestrian tracking using smartphones,'' \emph{IEEE
  Sensors Journal}, vol.~16, no.~7, pp. 2079--2093, 2015.

\bibitem{blackmanP:1999}
S.~S. Blackman and R.~F. Popoli, \emph{Design and Analysis of Modern Tracking
  Systems}.\hskip 1em plus 0.5em minus 0.4em\relax Artech House, 1999.

\end{thebibliography}

\end{document}